\documentclass[final]{cvpr}

\usepackage{times}
\usepackage{epsfig}
\usepackage{graphicx}
\usepackage{amsmath}
\usepackage{amssymb}
\usepackage{subcaption}
\usepackage{graphicx}
\usepackage{float}
\usepackage{algpseudocode,algorithm}
\usepackage{gensymb}
\usepackage{booktabs}
\usepackage{multirow}
\usepackage{stfloats}

\usepackage[pagebackref=true,breaklinks=true,colorlinks,bookmarks=false]{hyperref}

\def\cvprPaperID{2536} 
\def\confYear{CVPR 2021}

\begin{document}

\title{AdCo: Adversarial Contrast for Efficient Learning of Unsupervised Representations from Self-Trained Negative Adversaries}

\author{Qianjiang Hu$^1$\thanks{Q. Hu and X. Wang made an equal contribution to performing experiments while being mentored by G.-J. Qi.}, Xiao Wang$^{2*}$, Wei Hu$^1$, Guo-Jun Qi$^{3}$\thanks{Corresponding author: G.-J. Qi (Email: guojunq@gmail.com), who conceived the idea, formulated the method, and wrote the paper.}\vspace{1.5mm}\\
$^1$Peking University, $^2$Purdue University\\
$^3$Laboratory for MAchine Perception and LEarning (MAPLE)\\
{\url{http://maple-lab.net/}}
}

\maketitle

\begin{abstract}
\vspace{-4mm}
Contrastive learning relies on constructing a collection of negative examples that are sufficiently hard to discriminate against positive queries when their representations are self-trained. Existing contrastive learning methods either maintain a queue of negative samples over minibatches while only a small portion of them are updated in an iteration, or only use the other examples from the current minibatch as negatives. They could not closely track the change of the learned representation over iterations by updating the entire queue as a whole, or discard the useful information from the past minibatches. Alternatively, we present to directly learn a set of negative adversaries playing against the self-trained representation. Two players, the representation network and negative adversaries, are alternately updated to obtain the most challenging negative examples against which the representation of positive queries will be trained to discriminate. We further show that the negative adversaries are updated towards a weighted combination of positive queries by maximizing the adversarial contrastive loss, thereby allowing them to closely track the change of representations over time.  Experiment results demonstrate the proposed Adversarial Contrastive (AdCo) model not only achieves superior performances (a top-1 accuracy of 73.2\% over 200 epochs and 75.7\% over 800 epochs with linear evaluation on ImageNet), but also can be pre-trained more efficiently with much shorter GPU time and fewer epochs. 
The source code is available at \url{https://github.com/maple-research-lab/AdCo}.
\end{abstract}

\vspace{-0.5\baselineskip}
\section{Introduction}
\vspace{-0.5\baselineskip}
Learning visual representations in an unsupervised fashion \cite{agrawal2015learning,qi2019learning,doersch2015unsupervised,kim2018learning,larsson2016learning,zhang2019aet} has attracted many attentions as it greatly reduces the cost of collecting a large volume of labeled data to train deep networks. Significant progresses have also been made to reduce the performance gap with the fully supervised models. Among them are a family of contrastive learning methods \cite{he2020momentum,chen2020simple,wu2018unsupervised,hjelm2018learning,oord2018representation,bachman2019learning,zhuang2019local,tian2019contrastive,henaff2019data} that self-train deep networks by distinguishing the representation of positive queries from their negative counterparts. Depending on how the negative samples are constructed, two large types of contrastive learning approaches have been proposed in literature \cite{he2020momentum,chen2020simple}. These negative samples play a critical role in contrastive learning since the success in self-training deep networks relies on how positive queries can be effectively distinguished from negative examples.

Specifically, one type of contrastive learning methods explicitly maintains a queue of negative examples from the past minibatches.  For example, the Momentum Contrast (MoCo) \cite{he2020momentum} iteratively updates an underlying queue with the representations from the current minibatch in a First-In-First-Out (FIFO) fashion. However, only a small portion of oldest negative samples in the queue would be updated, which could not continuously track the rapid change of the feature representations over iterations.  Even worse, the momentum update of key encoders, which is necessary to stabilize the negative queue in MoCo, could further slow down the track of the representations. Consequently, this would inefficiently train the representation network, since partially updated negatives may not cover all critically challenging samples thus far that ought to be distinguished from positive queries to train the network.

Alternatively, another type of contrastive learning methods \cite{chen2020simple} abandons the use of such a separate queue of negative examples.  Instead, all negative examples come from the current minibatch, and a positive query would be retrieved by distinguishing it from the other examples in the minibatch. However, it discards the negative examples from the past minibatches, and often requires a much larger size of minibatch so that a sufficient number of negative samples are available to train the representation network by discriminating against positive queries. This incurs heavy memory and computing burden to train over each minibatch for this type of contrastive learning.

In this paper, we are motivated to address the aforementioned drawbacks, and the objective is twofold. First, we wish to construct a set of negative examples that can continuously track the change of the learned representation rather than updating only a small portion of them.  In particular, it will update negative samples {\em as a whole} by making them sufficiently  challenging to train a representation network more efficiently with fewer epochs.  On the other hand, it will retain the discriminative information from the past iterations without depending on a much larger size of minibatch to train the network \cite{chen2020simple}. We will show that in the proposed model, the negative examples are directly {\em trainable} so that they can be integrated as a part of the underlying network and trained end-to-end together with the representation. Thus, the trainable negatives are analogous to the additional network component in other self-supervised models without involving negative examples, such as the prediction MLP of the BYOL \cite{grill2020bootstrap} that also needs to be trained end-to-end. More discussions on {\em whether we still need negative examples} can be found in Appendix~\ref{appendix:nsamples}.

Particularly, we will present an Adversarial Contrast (AdCo) model consisting of two adversarial players. One is a backbone representation network that encodes the representation of input samples. The other is a collection of negative adversaries that are used to discriminate against positive queries over a minibatch.  Two players are alternately updated. With a fixed set of negative adversaries, the network backbone is trained by minimizing the contrastive loss of mistakenly assigning positive queries to negative samples as in the conventional contrastive learning.  On the other hand, the negative adversaries are updated by {\em maximizing} the contrastive loss, which pushes the negative samples to closely track the positive queries over the current minibatch. This results in a minimax problem to find an equilibrium as its saddle point solution. Although there is no theoretical guarantee of convergence, iterative gradient updates to the network backbone and the negative adversaries work well in practice, which has been observed in many other adversarial methods \cite{goodfellow2014generative,qi2020loss,qi2018global,zhao2018adversarial}. We will also show that the derivative of the contrastive loss wrt the negative adversaries reveals how they are updated towards a weighted combination of positive queries, and gives us an insight into how the AdCo focuses on low-density queries compared with an alternative model.


The experiment results not only demonstrate the AdCo has the superior performance on downstream tasks, but also verify that it can train the unsupervised networks with fewer epochs by updating the negative adversaries more efficiently.
For example, with merely $10$ epochs of pretraining, the AdCo has a top-1 accuracy of $44.4\%$, which is almost $5\%$ higher than that of the MoCo v2 pretrained over the same number of epochs with the linear evaluation on the ResNet-50 backbone on ImageNet. It also greatly outperforms the MoCHi \cite{kalantidis2020hard} by $4.1\%$ in top-1 accuracy over 800 epochs that enhances the MoCo v2 with mixed hard negatives, showing its effectiveness in constructing more challenging negative adversaries in a principled fashion to pretrain the representation network.

Moreover, the AdCo achieves a record top-1 accuracy of $75.7\%$ over $800$ epochs compared with the state-of-the-art BYOL (74.3\%) and SWAV (75.3\%) models pretrained for $800\sim 1,000$ epochs. This is obtained with the same or even smaller amount of GPU time than the two top-performing models.
Indeed, the AdCo is computationally efficient with a negligible cost of updating negative adversaries, making it an attractive paradigm of contrast model having higher accuracies over fewer epochs with no increase in the computing cost.

The remainder of the paper is organized as follows. We will review the related works in Section~\ref{sec:related}, and present the proposed approach in Section~\ref{sec:approach}.  By comparing the proposed AdCo with an alternative form of adversarial contrastive loss, we will reveal how AdCo focuses on low-density queries whose representations have not been well captured in Section~\ref{sec:disc}. We will conduct experiments in Section~\ref{sec:exp} to demonstrate its superior performance in multiple tasks. Finally, we will conclude in Section~\ref{sec:concl}.

\vspace{-0.5\baselineskip}
\section{Related Works}\label{sec:related}
\vspace{-0.5\baselineskip}

In this section, we review the related works on unsupervised representation learning. In particular, we will review two large families of unsupervised learning methods: the contrastive learning that is directly related with the proposed Adversarial Contrastive Learning (AdCo), and the alternative approach that instead aims to learns the transformation equivariant features without labeled data. We will see that while the former explores the {\em inter-instance discrimination} to self-supervise the training of deep networks, the latter leverages {\em intra-instance variation} to learn transformation-equivariant representations.

\subsection{Contrastive Learning}
Originally, contrastive learning \cite{oord2018representation,henaff2019data} was proposed to learn unsupervised representations by maximizing the mutual information between the learned representation and a particular context \cite{bachman2019learning,zhuang2019local,tian2019contrastive}. It usually uses the context of the same instance to learn representations by discriminating between positive queries and a collection of negative examples in an embedding feature space \cite{wu2018unsupervised,he2019momentum,chen2020simple}. Among them is the instance discrimination that has been used as a pretext task by distinguishing augmented samples from each other in a minimatch \cite{chen2020simple}, over a memory bank \cite{wu2018unsupervised}, or a dynamic queue \cite{he2019momentum} with a momentum update.  The retrieval of a given query is usually performed by matching it against an augmentation of the same sample from a separate collection of negative examples.  These negative examples play a critical role to challenge the self-trained representation so that it can become gradually more discriminative over iterations to distinguish the presented queries from those hard negative examples. Recently, Grill et al.~\cite{grill2020bootstrap} propose to train deep networks by predicting the representation of an image from that of an different augmented view of the same image. Kalantidis et al.~\cite{kalantidis2020hard} present a feature-level sample mixing strategies to generate harder meaningful negatives to further improve network pretraining. However, this differs from the proposed AdCo that aims to update negative examples in a more principle way through an adversarial self-training mechanism.

\subsection{Other Unsupervised Methods}

An alternative approach to unsupervised representation learning is based on transformation prediction \cite{zhang2019aet,gidaris2018unsupervised}. On contrary to contrastive learning that focuses on inter-instance discrimination, it attempts to learn the representations equivarying to various transformations on 2D plenary images and 3D cloud points \cite{qi2019avt,qi2019learning,gao2020graphter}.  These transformations are applied to augment training examples and learn their representations from which the transformations can be predicted from the captured visual structures.  It focuses on modeling the {\em intra-instance variations} from which transformation-equivariant representations can be leveraged on downstream tasks such as image classification \cite{zhang2019aet,qi2019learning,gidaris2018unsupervised}, object detection \cite{gidaris2018unsupervised,qi2019learning}, semantic segmentations \cite{gidaris2018unsupervised,qi2016hierarchically} and 3D cloud points \cite{gao2020graphter}.  This category of methods can be viewed as orthogonal to contrastive learning approaches that are based on {\em inter-instance discrimination}. More comprehensive review of recent advances on unsupervised methods can be found in \cite{qi2019small}.

\vspace{-0.5\baselineskip}
\section{The Proposed Approach}\label{sec:approach}
\vspace{-0.5\baselineskip}
In this section, we will first briefly review the preliminary work on contrastive learning and discuss its drawbacks that motivate the proposed approach in Section~\ref{sec:pre}.  Then, we will present the proposed method in Section~\ref{sec:joint} to jointly train the representation network and negative adversaries.  The insight into the proposed approach can be better revealed from the derivative of the adversarial contrastive loss leading to the update rule for the negative samples in Section~\ref{sec:insight}.

\subsection{Preliminaries and Motivations}\label{sec:pre}

We begin by briefly revisiting contrastive learning and its variants, and discuss their limitations that motivate the proposed work. In a typical contrastive learning method, we seek to learn an unsupervised representation by minimizing a contrastive loss $\mathcal L$. Specifically, in a minibatch $\mathcal B$ of $N$ samples, consider a given query $\mathbf x_i$ and the embedding $\mathbf q_i$ of its augmentation through a backbone network with the parametric weights $\theta$. The contrastive learning aims to train the network $\theta$ such that the query $q$ can be distinguished from a set $\mathcal N=\{\mathbf n_k|k=1,\cdots,K\}$ of the representations of negative samples. Formally, the following contrastive loss is presented in InfoNCE \cite{oord2018representation} in a soft-max form,
\begin{equation}\label{eq:closs}
\small
\mathcal L(\theta) = \dfrac{-1}{N} \sum_{i=1}^N \log \dfrac{\exp(\mathbf q_i^\intercal \mathbf q'_i/\tau)}{\exp(\mathbf q_i^\intercal \mathbf q'_i/\tau)+\sum_{k=1}^K\exp(\mathbf q_i^\intercal \mathbf n_k/\tau)}
\end{equation}
where $\mathbf q'_i$ is the embedding of another augmentation of the same instance $\mathbf x_i$, which is considered as the positive example for the query $\mathbf q_i$; and $\tau$ is a positive value of temperature. Then the network can be updated in each minibatch by minimizing this loss over $\theta$.  Note that the representations of both queries and negative samples are $\ell_2$ normalized such that the dot product results in a cosine similarity between them.

There are two main categories of methods about how to construct the set of negative samples $\mathcal N$ in contrast to positive queries. These negative samples play a critical role in self-training unsupervised representations, and deserve a careful investigation. The first category of contrastive learning approaches maintain a queue of negative representations that are iteratively updated in a First-In-First-Out (FIFO) fashion over minibatches \cite{he2019momentum}, while the alternative type of approaches only adopt the other samples as negatives from the current minibatch \cite{chen2020simple}.


However, both types suffer some drawbacks. For the first type of approaches, in each iteration over a minibatch, only a small portion of negative representations in the queue are updated, and this results in an under-represented set that fails to cover the most critical negative samples to train the representation network, since not all of them have been updated timely to track the change of unsupervised representations.  On the other hand, the other type completely abandons the negative samples from the past minibatches, which forces them to adopt a much larger size of minibatch to construct a sufficient number of negative representations. Moreover, only using the other samples from the current minibatch as negatives discards rich information from the past.

To address these drawbacks, we propose to {\em actively} train a set of negative examples as a whole in an adversarial fashion, in contrast to passively queueing the negative samples over minibatches or simply using the other samples from the current minibatch as negatives.  Its advantages are twofold. First,  the entire set rather than only a small portion of negative representations will be updated as a whole. This enables the learned negative adversaries to not only closely track the change of the learned representation but also keep the accumulative information from the past minibatches.

Moreover, the set of negative samples will act as adversaries to the representation network $\theta$ by {\em maximizing} the contrastive loss, which will result in a minimax problem to self-train the network.  As known in literature, a self-supervisory objective ought to be sufficiently challenging to prevent the learned representation from overfitting to the objective or finding a bypassing solution that could be trivial to downstream tasks.  In this sense, the adversarially learned negative set can improve the generalizability of the learned representation as an explicit adversary to challenge the updated network by engaging the most challenging negative samples over epochs.

\subsection{Adversarial Training of Representation Networks and Negative Samples}\label{sec:joint}
In this section, we formally present the proposed Adversarial Contrastive Learning (AdCo) of unsupervised representations. It consists of two mutually interacted players: a deep representation network $\theta$ embedding input examples into feature vectors, and a set $\mathcal N$ of negative adversaries closely tracking the change of the learned representation. While negative adversaries contain the most critical examples that are supposed to be confused with given queries to challenge the representation network, the network is self-trained to continuously improve its discrimination ability to distinguish positive queries from those challenging negative adversaries. Eventually, we hope an equilibrium can be reached where the learned representation can achieve a maximized performance.


Formally, this results in the following minimax problem to train both players jointly with the adversarial contrastive loss $\mathcal L$ as in (\ref{eq:closs}),
\begin{equation}\label{eq:adcl}
\theta^\star,\mathcal N^\star = \arg\min_\theta\max_{\mathcal N} \mathcal L(\theta,\mathcal N)
\end{equation}
where the embedding $\mathbf q_i$ of each query in the current minibatch are a function of network weights $\theta$, and thus we optimize over $\theta$ through them; the negative adversaries in $\mathcal N$ are treated as free variables directly, which are unit-norm vectors subject to $\ell_2$-normalization.

As in many existing adversarial training algorithms, it is hard to find a saddle point $(\theta^\star,\mathcal N^\star)$ solution to the above minimax problem. Usually, a pair of gradient descent and ascent are applied to update them, respectively
\begin{equation}\label{eq:theta}
\theta \longleftarrow \theta - \eta_\theta \dfrac{\partial\mathcal L(\theta,\mathcal N)}{\partial \theta}
\end{equation}
\begin{equation}\label{eq:N}
\mathbf n_k \longleftarrow \mathbf n_k + \eta_{\mathcal N} \dfrac{\partial\mathcal L(\theta,\mathcal N)}{\partial \mathbf n_k}\footnote{Strictly speaking, the update of $\mathbf n_k$ ought to be performed by taking the derivative of $\mathcal L$ wrt the one prior to $\ell_2$-normalized. However, to ease the exposition later, we simply adopt the derivative wrt the normalized embedding without the loss of generality.}
\end{equation}
for $k=1,\cdots,K$, where $\eta_\theta$ and $\eta_{\mathcal N}$ are the positive learning rates for updating the network and negative adversaries. Although no theory guarantees the convergence to the saddle point, it works well in our experiments by alternately updating $\theta$ and $\mathcal N$.

Before we take an insight into the updated $\mathbf n_k$'s, an intuitive explanation can be given by noticing that the cosine similarities between the negative samples $\mathbf n_k$'s and the queries $\mathbf q_i$ in the denominator of (\ref{eq:closs}) are maximized when $\mathcal L(\theta, \mathcal N)$ is maximized for the adversarial training of $\mathcal N$. In other words, this tends to push the negative samples closer towards the queries from the current minibatch, thus resulting in challenging negatives closely tracking the change of the updated network.

\subsection{Derivatives of The AdCo Loss}\label{sec:insight}

Let us look into the update rule (\ref{eq:N}) for the negative samples $\mathbf n_k, k= 1, \cdots, K$.
It is not hard to show that the derivative of the adversarial loss in updating a negative sample $\mathbf n_k$ is
\begin{equation}\label{eq:dloss}
\small
\dfrac{\partial\mathcal L}{\partial \mathbf n_k} = \frac{1}{N\tau}\sum_{i=1}^N \dfrac{\exp(\mathbf q_i^\intercal \mathbf n_k/\tau) \cdot \mathbf q_i }{\exp(\mathbf q_i^\intercal \mathbf q'_i/\tau)+\sum_{k=1}^K\exp(\mathbf q_i^\intercal \mathbf n_k/\tau)}
\end{equation}
where the first factor can be viewed as the conditional probability of assigning the query $\mathbf q_i$ to a negative sample $\mathbf n_k$,
$$
p(\mathbf n_k|\mathbf q_i) \triangleq \dfrac{\exp(\mathbf q_i^\intercal \mathbf n_k/\tau) }{\exp(\mathbf q_i^\intercal \mathbf q'_i/\tau)+\sum_{k=1}^K\exp(\mathbf q_i^\intercal \mathbf n_k/\tau)}
$$
This is a valid probability because it is always nonnegative and the sum of all such conditional probabilities is one, i.e.,
$$
p(\mathbf q'_i|\mathbf q_i) + \sum_{k=1}^K p(\mathbf n_k|\mathbf q_i)=1.
$$

Now the derivative (\ref{eq:dloss}) can be rewritten as
\begin{equation}\label{eq:weighted}
\dfrac{\partial\mathcal L}{\partial \mathbf n_k} = \dfrac{1}{N\tau}\sum_{i=1}^N p(\mathbf n_k|\mathbf q_i)~\mathbf q_i.
\end{equation}

This reveals the physical meaning by updating each negative sample $\mathbf n_k$ along the direction given by a $p(\mathbf n_k|\mathbf q_i)$-weighted combination of all the queries $\mathbf q_i$ over the current minibatch. The more likely a negative sample is assigned to a query, the closer the sample is pushed towards the query. This will force the negative samples to continuously track the queries that are most difficult to distinguish, and thus give rise to a challenging set of adversarial negative samples to more critically train the underlying representation. Moreover, this adversarial update also tends to cover low-density queries that have not been well captured by the current set of negative samples, which we will discuss in the next section.
\vspace{-0.5\baselineskip}
\section{Further Discussions}\label{sec:disc}
\vspace{-0.5\baselineskip}
In this section, first we will present an alternative of the proposed AdCo in Section~\ref{sec:ext}. Then, based on it, we will review the derivative of the AdCo loss in Section~\ref{sec:adapt}, and show that it seeks to adapt negative adversaries to low-density queries, thereby allowing more efficient adversarial training of the representation network.

\subsection{An Alternative of AdCo}\label{sec:ext}
An alternative adversarial loss can be motivated from (\ref{eq:weighted}). One can adopt another form of conditional probability as the weighting factor by constructing a new adversarial loss $\mathcal J(\theta,\mathcal N)$ to train the negative adversaries. In particular, we may consider the conditional probability $p(\mathbf q_i|\mathbf n_k)$ in place of $p(\mathbf n_k|\mathbf q_i)$ used in (\ref{eq:weighted}). This turns out to be a more natural choice since the derivative becomes the conditional expectation of the query $\mathbf q_i$'s, i.e.,
\begin{equation}\label{eq:mloss}
\dfrac{\partial\mathcal J(\theta,\mathcal N)}{\partial \mathbf n_k} = \sum_{i=1}^N p(\mathbf q_i|\mathbf n_k)~\mathbf q_i \triangleq \mathbb E_{p(\mathbf q_i|\mathbf n_k)} \left[\mathbf q_i|\mathbf n_k\right].
\end{equation}

In this case, it is not hard to verify that the corresponding adversarial loss $\mathcal J$ satisfying the above derivative relation can be defined as
$$
\max_{\mathcal N} \mathcal J(\theta,\mathcal N) \triangleq -\tau \cdot \sum_{k=1}^K \log \dfrac{1}{\sum_{i=1}^N\exp(\mathbf q_i^\intercal \mathbf n_k/\tau)}
$$
with the following conditional probability of assigning the negative sample $\mathbf n_k$ to a target query $\mathbf q_i$
$$
p(\mathbf q_i|\mathbf n_k) = \dfrac{\exp(\mathbf q_i^\intercal \mathbf n_k/\tau)}{\sum_{i=1}^N\exp(\mathbf q_i^\intercal \mathbf n_k/\tau)}
$$
Similarly, it is not hard to show that this is a valid conditional probability by verifying it is nonnegative and satisfies its sum being one. Then, the negative samples are updated by gradient ascent on $\mathcal J(\theta, \mathbf N)$ over $\mathbf n_k$'s.

\subsection{Adapting to Low-Density Queries}\label{sec:adapt}

Although the derivative of this adversarial loss $\mathcal J$ embodies a more explicit meaning as in (\ref{eq:mloss}), our preliminary experiments with this loss to update negative samples perform worse than that of the AdCo in (\ref{eq:adcl}). This probably is due to the fact that the loss $\mathcal J$ does not directly adverse against the contrastive loss $\mathcal L$ used to train the embedding network $\theta$.  Thus, the generated negative samples may not provide sufficiently hard negative samples as direct adversaries to the contrastive loss (\ref{eq:adcl}) used to train the representation network.

Indeed, by applying the Bayesian rule to $p(\mathbf n_k|\mathbf q_i)$, the derivative in (\ref{eq:weighted}) can be rewritten as
$$
\dfrac{\partial\mathcal L}{\partial \mathbf n_k} \propto \sum_{i=1}^N \dfrac{p(\mathbf q_i|\mathbf n_k)p(\mathbf n_k)}{p(\mathbf q_i)}~\mathbf q_i =p(\mathbf n_k)  \mathbb E_{p(\mathbf q_i|\mathbf n_k)} \left[\widetilde {\mathbf q}_i|\mathbf n_k\right]
$$
where $\widetilde {\mathbf q}_i \triangleq \frac{\mathbf q_i}{p(\mathbf q_i)}$ is the embedded query $\mathbf q_i$ normalized by $p(\mathbf q_i)$, and we put $p(\mathbf n_k)$ outside the conditional expectation by viewing $\mathbf n_k$ as a constant in the condition.

It is not hard to see that for a low-density query $\mathbf q_i$ with a smaller value of $p(\mathbf q_i)$, it has a larger value of the normalized query $\widetilde {\mathbf q}_i$. Thus, the resultant derivatives will push negative adversaries closer to such low-density queries that have not been well represented by the negative examples \footnote{From $p(\mathbf q_i)=\sum_{k=1}^K p(\mathbf q_i|\mathbf n_k)p(\mathbf n_k)$, a low-density query $\mathbf q_i$ should have a small value of $p(\mathbf q_i|\mathbf n_k)$ for high-density negative adversaries $\mathbf n_k$'s, which means the query should be far apart from the dense negative examples in the embedding space.}.  This will enable more efficient training of the representation network with the negative adversaries updated to cover these under-represented queries.

For this reason, we will still use the AdCo loss (\ref{eq:adcl}) to jointly train the negative adversaries and the network in our experiments.  However, the loss (\ref{eq:mloss}) derived from the conditional expectation provides an insight into some future direction to alterative forms of adversarial loss to train negative samples.
While this gives us an alternative explanation of the proposed method, it could have some theoretical value that deserves further study in future.

%

\section{Experiments}\label{sec:exp}

In this section, we conduct experiments on the proposed AdCo model and compare it with the other existing unsupervised methods.

\subsection{Training Details}

For the sake of fair and direct comparison with the existing models \cite{he2019momentum,chen2020simple}, we adopt the ResNet-50 as the backbone for unsupervised pretraining on ImageNet. The output feature map from the top ResNet-50 block is average-pooled and projects to a 128-D feature vector through two-layer MLP (2048-D hidden lyaer with the ReLU) \cite{chen2020simple}. The resultant vector is $\ell_2$ normalized to calculate the cosine similarity. We apply the same augmentation proposed in the SimCLR \cite{chen2020simple} and adopted by the other methods \cite{he2019momentum,chen2020improved} to augment the images over each minibatch. Although it was reported in literature that carefully tuning the augmentation may improve the performance, we do not adopt it for a fair comparison with the other unsupervised methods while avoiding over-tuning the image augmentation strategy on the dataset.

For the unsupervised pretraining on ImageNet, we use the SGD optimizer (\cite{bottou2010large}) with an initial learning rate of $0.03$ and $3.0$ for updating the backbone network and the negative adversaries, respectively, where a weight decay of $10^{-4}$ and a momentum of $0.9$ are also applied. Cosine scheduler~(\cite{loshchilov2016sgdr}) is used to gradually decay the learning rate. We also choose a lower temperate ($\tau=0.02$) to update the negative adversaries than that ($\tau=0.12$) used for updating the backbone network. This makes the updated negative adversaries sharper in distinguishing themselves from the positive images, and thus they will be nontrivially challenging examples in training the network.


The batch size is set to 256. When multiple GPU servers are used, the batch size will be multiplied by the number of servers by convention. The number of negative adversaries is set to $65,536$, which is the same as the queue length of negative examples in MoCo \cite{he2019momentum}.
All negative adversaries are also $\ell_2$-normalized in each iteration to have an unit norm after they are updated by the SGD optimizer. The backbone network is first initialized, and its output feature vectors over randomly drawn training images are used to initialize the set of negative examples. Once the initialization is done, both the representation network and negative adversaries will be alternately updated by the AdCo.

\subsection{Experiment Results}
After the backbone network is pretrained, the resultant network will be evaluated on several downstream tasks.

\subsubsection{Linear Classification on ImageNet}

\begin{table}
\begin{center}
\caption{\small Top-1 accuracy under the linear evaluation on ImageNet with the ResNet-50 backbone. The table compares the methods over $200$ epochs of pretraining.}
\label{tab:classification200}
\setlength{\tabcolsep}{1.5mm}{
\small
\begin{tabular}{lccc}
\toprule[1pt]
Method & Batch size&Top-1 \\
\midrule[1pt]
InstDisc \cite{wu2018unsupervised} & 256  & 58.5 \\
LocalAgg \cite{zhuang2019local} & 128  & 58.8 \\
MoCo \cite{he2020momentum} & 256 &60.8 \\
SimCLR \cite{chen2020simple} & 256  & 61.9 \\
CPC v2 \cite{henaff2019data} & 512  & 63.8 \\
PCL v2 \cite{li2020prototypical} & 256  & 67.6 \\
MoCo v2 \cite{chen2020improved} & 256  & 67.5 \\
MoCHi \cite{kalantidis2020hard} & 512 & 68.0 \\
PIC \cite{cao2020parametric} & 512 & 67.6 \\
SWAV* \cite{caron2020unsupervised} & 256  & 72.7 \\
\midrule
AdCo &256  & {\bf 68.6} \\
AdCo* &256 & {\bf 73.2} \\
\bottomrule[1pt]
\end{tabular}}
\end{center}
{$^*$ with multi-crop augmentations.}
\end{table}


\begin{table}
\begin{center}
\caption{\small Running time (in hours) for different methods. The last column shows the total GPU time per epoch.}
\label{tab:time}
{
\small
\begin{tabular}{lllc}
\toprule[1pt]
Method &Epoch& GPU & (GPU$\cdot$Time)/Epoch \\
\midrule[1pt]
{MoCo v2} \cite{chen2020improved} &200 & 8$\times$V100 & 2.12\\
BYOL \cite{grill2020bootstrap}& 1000 & 512$\times$TPU & 4.10 \\
{SWAV*} \cite{caron2020unsupervised} & 200 & 64$\times$V100 & 4.06 \\
\midrule
AdCo &200& 8$\times$V100 & 2.26 \\
AdCo* &200 & 8$\times$V100 & 4.39\\
\bottomrule[1pt]
\end{tabular}}
\end{center}
{$^*$ with multi-crop augmentations.}
\end{table}

First, we compare the proposed AdCo with the other methods in terms of top-1 accuracy.  A linear fully connected classifier is fine-tuned on top of the frozen 2048-D feature vector out of the pretrained ResNet-50 backbone. The linear layer is trained for $100$ epochs, with a learning rate of $10$ with the cosine learning rate decay and a batch size of $256$.

Table~\ref{tab:classification200} reports experiment results with the backbone network pretrained for $200$ epochs. We note that the
state-of-the-art SWAV model has applied multi-crop augmentations to pretrain the model. Thus for the sake of a fair comparison, in addition to the single-crop model, we also report the result by applying five crops (224x224, 192x192, 160x160, 128x128, and 96x96) over minibatch images during the pretraining. The result shows the proposed AdCo achieves the best performance among the compared models. It not only outperforms the SOTA SWAV model in the top-1 accuracy over 200 epochs, but also has a computing time on par with the top-performing contrast models in terms of GPU$\cdot$Time per epoch as shown in Table~\ref{tab:time}. It is not surprising since the computing overhead for updating the negative adversaries is negligible compared with that for updating the backbone network.

Note that some methods such as the BYOL \cite{grill2020bootstrap} and the SimCLR \cite{chen2020simple} used symmetric losses by swapping pairs of augmented images to improve the performance. For a fair comparison, we also symmetrize the AdCo loss, and show in Appendix~\ref{appendix:sloss} that it obtains a competitive top-1 accuracy of 70.6\% over 200 epochs of pre-training based only on single-crop augmentation with much shorter GPU time and a smaller batch size. We refer the readers to the appendix for more details. In Appendix~\ref{appendix:nsamples}, we also seek the answer to an emerging question -- {\em do we still need negative samples to pre-train deep networks while the BYOL does not?} .

\begin{table}
\begin{center}
\caption{\small Top-1 accuracy under the linear evaluation on ImageNet with the ResNet-50 backbone. The table compares the methods with more epochs of network pretraining.}
\label{tab:classification800}
\small
\setlength{\tabcolsep}{3.5mm}{
\begin{tabular}{llll}
\toprule[1pt]
Method &Epoch& Batch size&  Top-1 \\
\midrule[1pt]
Supervised &-&-& 76.5\\
\hline
BigBiGAN\cite{donahue2019large}&-& 2048 & 56.6 \\
SeLa\cite{asano2019self} &400& 256  & 61.5 \\
PIRL\cite{misra2020self}&800& 1024 & 63.6 \\
CMC\cite{tian2019contrastive} &240& 128  & 66.2\\
SimCLR\cite{chen2020simple} &800& 4096 & 69.3\\
PIC\cite{cao2020parametric} &1600& 512 &  70.8 \\
MoCo v2\cite{chen2020improved} &800& 256 &  71.1 \\
MoCHi \cite{kalantidis2020hard} & 800 & 512 & 68.7 \\
BYOL **\cite{grill2020bootstrap}&1000& 4096 & 74.3\\
{SWAV*}\cite{caron2020unsupervised} &800& 4096 & 75.3 \\
\midrule
AdCo &800&256& {\bf 72.8}\\
AdCo* &800&1024& {\bf 75.7}\\
\bottomrule[1pt]
\end{tabular}}
\end{center}
{$^*$ with multi-crop augmentations.\\
$^{**}$ fine-tuned with a batch size of $1024$ over 80 epochs.}
\end{table}

We also compare the results over more epochs in Table~\ref{tab:classification800}. While the AdCo* has the running time on par with the other top-2 best models (SWAV and BYOL) during the pretraining (see Table~\ref{tab:time}), it has achieved a record top-1 accuracy of $75.7\%$ with linear evaluation on the ImageNet after $800$ epochs of pretraining. Even without multi-crop augmentations, the AdCo also outperforms the top-performing MoCo v2 \cite{chen2020improved} (72.8\% vs. 71.1\%) and the variant MoCHi (72.8\% vs. 68.7\%) \cite{kalantidis2020hard} that enhances the MoCo v2 with mixed hard negative examples. This demonstrates that the AdCo is much more effective in constructing challenging negative examples to pre-train the representation network than not only the typical contrast model like the MoCo v2 but also its variant explicitly mining hard samples.


\subsubsection{Transfer Learning and Object Detection}


\begin{table*}
\begin{center}
\caption{Transfer learning results on cross-dataset classification and object detection tasks.}\label{tab:downstream}
\small
\setlength{\tabcolsep}{7mm}{
\begin{tabular}{lcccccc}
\toprule[1pt]
 & &\multicolumn{2}{c} {Classification} & \multicolumn{3}{c}{Object Detection} \\
 \cmidrule(lr){3-4}
 \cmidrule(lr){5-7}
 &  &VOC07 & Places205 & VOC07+12 & \multicolumn{2}{c}{COCO} \\
Method & Epoch  & mAP & Top-1 & $AP_{50}$ & $AP$ & $AP_{S}$ \\
\midrule[1pt]
Supervised & - & 87.5 &53.2 & 81.3 & 40.8 & 20.1 \\ \hline
NPID++~\cite{wu2018unsupervised} & 200 & 76.6 & 46.4 & 79.1 & - & - \\
MoCo~\cite{he2020momentum} & 800  & 79.8 & 46.9 & 81.5 & - & - \\
PIRL~\cite{misra2020self} & 800 & 81.1 & 49.8 & 80.7 & - & - \\
PCLv2~\cite{li2020prototypical} & 200 & 85.4 & 50.3 & 78.5 & - & - \\
BoWNet~\cite{gidaris2020learning} & 280  & 79.3 & 51.1 & 81.3 & - & - \\
SimCLR~\cite{chen2020simple} & 800 & 86.4 & 53.3 & - & - & - \\
MoCo v2~\cite{chen2020improved} & 800  & 87.1 & 52.9 & 82.5 & {\bf 42.0} & 20.8 \\
SWAV*~\cite{caron2020unsupervised} & 800  & 88.9 & {\bf 56.7} & 82.6 & 42.1 & 19.7 \\
\midrule
AdCo & 800 & {\bf 93.1} & {\bf 53.7} & {\bf 83.1} &41.8  & {\bf 24.1}  \\
AdCo* & 800  & {\bf 92.4} & 56.0 & {\bf 83.0} & {\bf 42.2} &  {\bf 24.5} \\
\bottomrule[1pt]
\end{tabular}}
\end{center}
{$^*$ with multi-crop augmentations.}
\end{table*}

We also evaluate the ResNet-50 pretrained on ImageNet for two downstream tasks -- cross-dataset transfer learning and object detection.  For the transfer learning on the VOC dataset \cite{everingham2010pascal}, we keep the pretrained ResNet-50 backbone frozen, fine-tune the linear classifier on VOC07trainval, and test on the VOC07test. We also train a linear classifier with the pretrained ResNet-50 on the Places dataset by following the previous evaluation protocol in literature \cite{chen2020simple,chen2020improved,caron2020unsupervised}.

For object detection, we adopt the same protocol in \cite{he2020momentum} to fine-tune the pretrained ResNet-50 backbone based on detectron2 \cite{wu2019detectron2} for a direct comparison with the other methods. On the VOC dataset \cite{everingham2010pascal}, the detection network is fine-tuned with VOC07+12 trainval dataset and tested on VOC07 test set. On the COCO dataset \cite{lin2014microsoft}, the detection network is fine-tuned on the train2017 dataset with 118k images, and is evaluated on the val2017 dataset.

Table~\ref{tab:downstream} shows the proposed AdCo achieves much competitive results on both tasks, suggesting the AdCo has better generalizability to the downstream tasks than many compared methods. For example, it has achieved a much higher accuracy of 93.1\% for the VOC07 classification task than the SWAV (88.9\%) based on the linear evaluation on the pretrained network, even without multi-crop augmentations. Moreover, the COCO dataset is well known for its challenging small object detection task measured by $AP_S$, and the result shows that the AdCo can significantly improve the $AP_S$  by $3.3\%-3.6\%$ no matter if multi-crop augmentations are applied.  This is a striking result as the SWAV with the multi-crop augmentations is even worse than the MoCo v2. This implies that the negative adversaries constructed by the AdCo may correspond to small objects, which in turn pushes the associated network to learn the representation highly discriminative in recognizing these challenging objects.


\subsection{Analysis and Visualization of Results}

\begin{figure}[t]
    \centering
    \begin{subfigure}[c]{0.4\textwidth}
        \includegraphics[width=\textwidth]{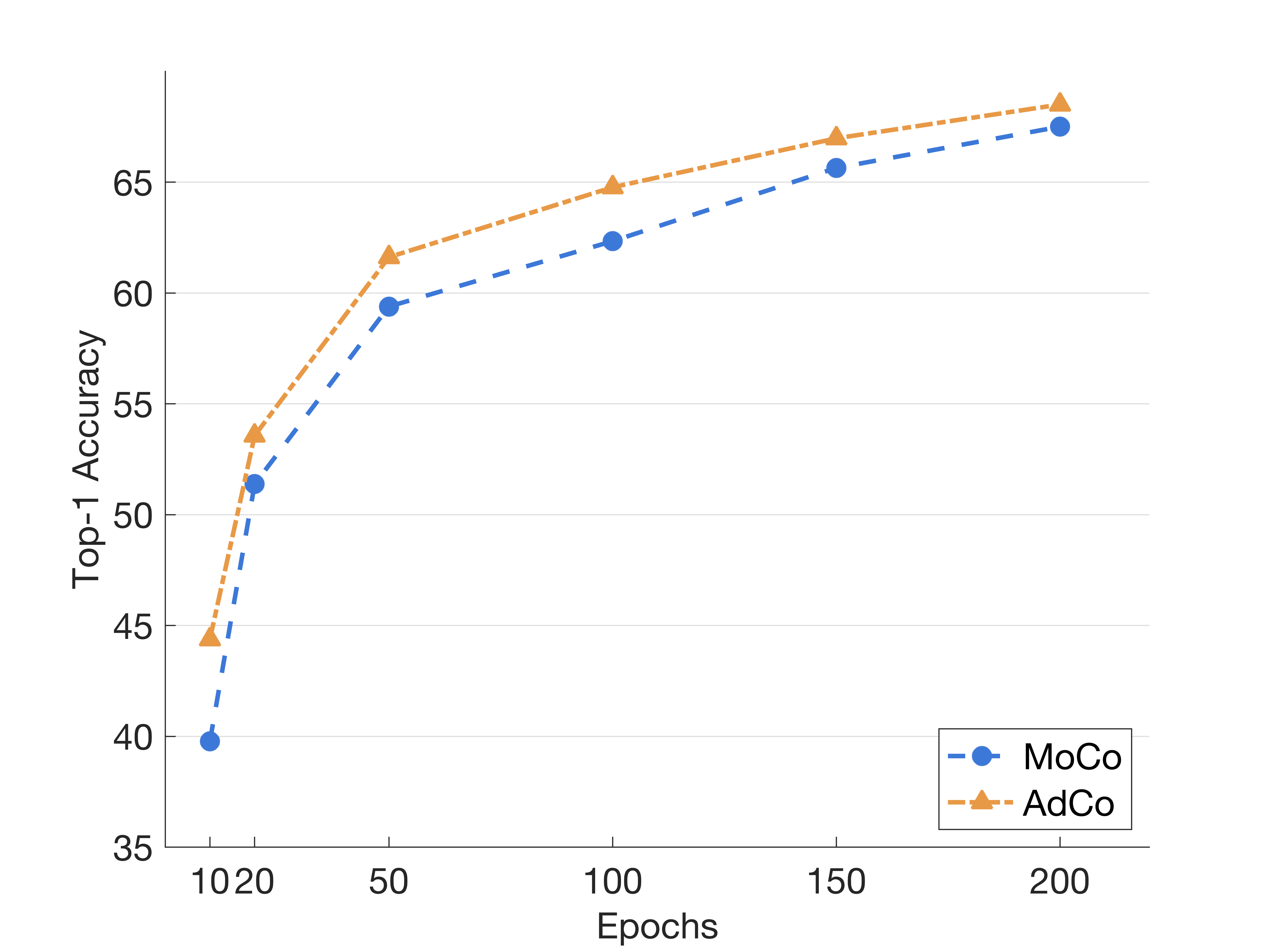}
    \end{subfigure}
    \caption{\small Top-1 accuracy of AdCo and MoCo V2 on ImageNet fine-tuned with a single fully connected layer upon the backbone ResNet-50 networks pretrained over up to $200$ epochs.  For a fair comparison, the AdCo maintains the same size ($K=65,536$) of negative samples as in MoCo v2 so that both models are directly compared under various numbers of epochs. With an extremely small number of $10$ epochs, the AdCo performs much better than the MoCo v2 by more than $5\%$ in top-1 accuracy. As the epoch increases, the AdCo usually achieves the performance comparable to that of the MoCo v2 with about $30 \sim 50$ fewer epochs. This shows the AdCo is more efficient in constructing a more critical collection of negative adversaries to improve network pretraining.}\label{fig:AdMoCo}
\end{figure}

\begin{figure*}[t]
    \centering
    \begin{subfigure}[c]{0.4\textwidth}
        \includegraphics[width=\textwidth]{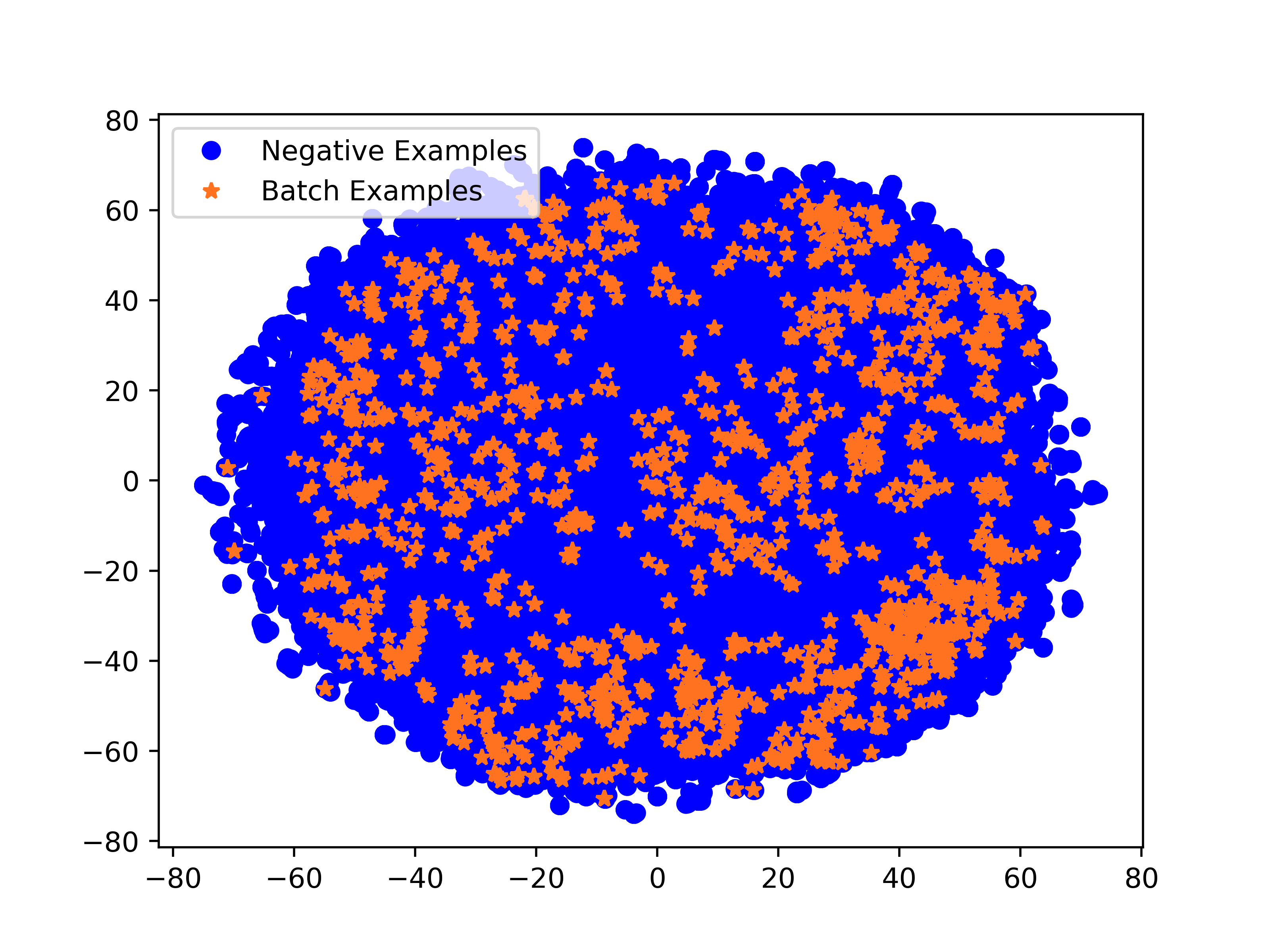}
        \caption{AdCo}
    \end{subfigure}\hspace{10mm}
        \begin{subfigure}[c]{0.4\textwidth}
        \includegraphics[width=\textwidth]{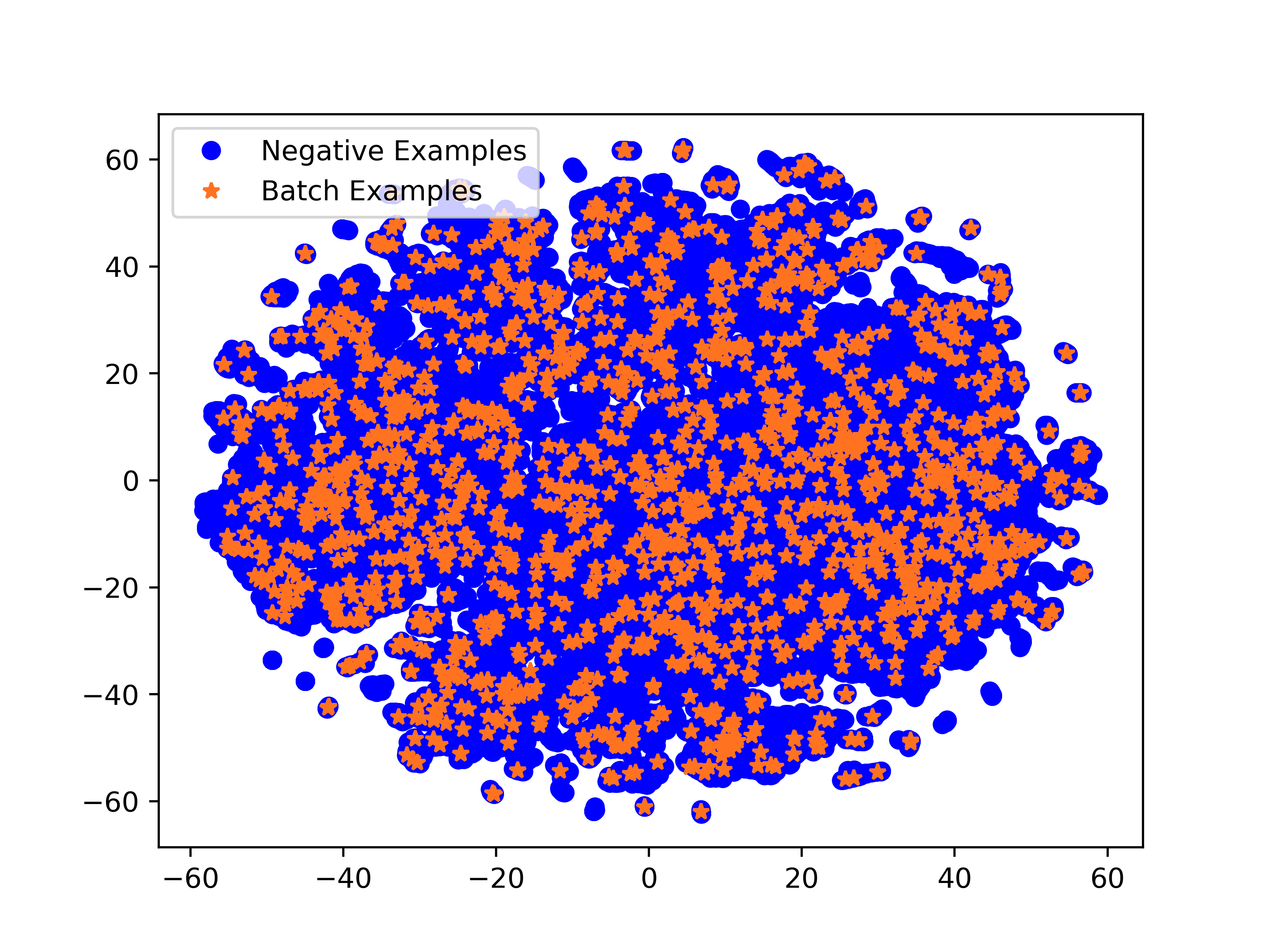}
        \caption{MoCo v2}
    \end{subfigure}
    \caption{t-SNE visualization of negative representations obtained by the AdCo and the MoCo in the 2D representation plane, alongside positive examples from the most recent six minibatches after $200$ epochs of pretraininig on the ImageNet. The figure shows that the AdCo has fewer outliers of negative examples than the MoCo v2, and thus more closely tracks the representation of positive samples over epochs. }\label{fig:negex}
\end{figure*}

First, we perform a comparative study of how the top-1 accuracy of AdCo and MoCo v2 changes over epochs.  As the state-of-the-art contrast model, the MoCo v2 also maintains a set of negative examples that play a critical role to train the representation network. As illustrated in Figure~\ref{fig:AdMoCo}, we plot the curve of their top-1 accuracies on ImageNet with a linear evaluation on the pretrained backbone under various numbers of epochs. The same number of $K=65,536$ negative examples are used in both models to ensure a fair comparison.

With an extreme small number of epochs, the result shows that the AdCo greatly outperforms the MoCo v2 by a significant margin. With $10$ epochs the top-1 accuracy of the AdCo is $5\%$ higher than that of the MoCo v2. With more epochs, the AdCo can reach the same level of accuracy with about $30 \sim 50$ fewer epochs than the MoCo v2.   This shows the AdCo can serve as an efficient paradigm of contrast model with fewer epochs of pretraining that does not increasing the computing cost as shown in Table~\ref{tab:time}.

We also compare the obtained negative examples by the AdCo and MoCo v2 by plotting t-SNE visualization in Figure~\ref{fig:negex}. We note that the MoCo v2 has more outliers of negative examples than the AdCo. These negative outliers form many small clusters that are isolated from most of batch examples in the learned representation space. Thus, they have little contributions to the contrastive training of the representation network as it is much easier to distinguish these negative outliers from positive batch examples.
On the contrary, with fewer negative outliers, the AdCo can more closely track the representation of batch examples over iterations, and thus efficiently train the representation network with more challenging negative adversaries.

More experiment results on the impact of various model designations and hyperparameters are presented in the appendix of this paper.




\section{Conclusions}\label{sec:concl}
This paper presents an Adversarial Contrast (AdCo) model by learning challenging negative adversaries that can be used to criticize and further improve the representation of deep networks. Compared with existing methods accumulating negative examples over the past minibatches and the other queries from the current minibatch, these negative adversaries in the AdCo are obtained by {\em maximizing} the adversarial contrastive loss of mis-assigning each positive query to negative samples. This updates the negative examples as a whole to closely track the changing representation, thus making it more challenging to distinguish them from positive queries. Consequently, the representation network must be efficiently updated to produce more discriminate representation. By analyzing the derivative of the adversarial objective, we show that each negative adversary is pushed towards a weighted combination of positive queries. Experiment results on multiple downstream tasks demonstrate its superior performances and the efficiency in pretraining the representation network.

{\small
\bibliographystyle{ieee_fullname}
\bibliography{unsupervised,AdCo}

\begin{thebibliography}{10}\itemsep=-1pt

\bibitem{agrawal2015learning}
Pulkit Agrawal, Joao Carreira, and Jitendra Malik.
\newblock Learning to see by moving.
\newblock In {\em Proceedings of the IEEE international conference on computer
  vision}, pages 37--45, 2015.

\bibitem{asano2019self}
Yuki~Markus Asano, Christian Rupprecht, and Andrea Vedaldi.
\newblock Self-labelling via simultaneous clustering and representation
  learning.
\newblock {\em arXiv preprint arXiv:1911.05371}, 2019.

\bibitem{bachman2019learning}
Philip Bachman, R~Devon Hjelm, and William Buchwalter.
\newblock Learning representations by maximizing mutual information across
  views.
\newblock In {\em Advances in Neural Information Processing Systems}, pages
  15509--15519, 2019.

\bibitem{bottou2010large}
L{\'e}on Bottou.
\newblock Large-scale machine learning with stochastic gradient descent.
\newblock In {\em Proceedings of COMPSTAT'2010}, pages 177--186. Springer,
  2010.

\bibitem{cao2020parametric}
Yue Cao, Zhenda Xie, Bin Liu, Yutong Lin, Zheng Zhang, and Han Hu.
\newblock Parametric instance classification for unsupervised visual feature
  learning.
\newblock {\em Advances in Neural Information Processing Systems}, 33, 2020.

\bibitem{caron2020unsupervised}
Mathilde Caron, Ishan Misra, Julien Mairal, Priya Goyal, Piotr Bojanowski, and
  Armand Joulin.
\newblock Unsupervised learning of visual features by contrasting cluster
  assignments.
\newblock {\em arXiv preprint arXiv:2006.09882}, 2020.

\bibitem{chen2020simple}
Ting Chen, Simon Kornblith, Mohammad Norouzi, and Geoffrey Hinton.
\newblock A simple framework for contrastive learning of visual
  representations.
\newblock {\em arXiv preprint arXiv:2002.05709}, 2020.

\bibitem{chen2020improved}
Xinlei Chen, Haoqi Fan, Ross Girshick, and Kaiming He.
\newblock Improved baselines with momentum contrastive learning.
\newblock {\em arXiv preprint arXiv:2003.04297}, 2020.

\bibitem{chen2020exploring}
Xinlei Chen and Kaiming He.
\newblock Exploring simple siamese representation learning.
\newblock {\em arXiv preprint arXiv:2011.10566}, 2020.

\bibitem{doersch2015unsupervised}
Carl Doersch, Abhinav Gupta, and Alexei~A Efros.
\newblock Unsupervised visual representation learning by context prediction.
\newblock In {\em Proceedings of the IEEE international conference on computer
  vision}, pages 1422--1430, 2015.

\bibitem{donahue2019large}
Jeff Donahue and Karen Simonyan.
\newblock Large scale adversarial representation learning.
\newblock In {\em Advances in Neural Information Processing Systems}, pages
  10542--10552, 2019.

\bibitem{everingham2010pascal}
Mark Everingham, Luc Van~Gool, Christopher~KI Williams, John Winn, and Andrew
  Zisserman.
\newblock The pascal visual object classes (voc) challenge.
\newblock {\em International journal of computer vision}, 88(2):303--338, 2010.

\bibitem{gao2020graphter}
Xiang Gao, Wei Hu, and Guo-Jun Qi.
\newblock Graphter: Unsupervised learning of graph transformation equivariant
  representations via auto-encoding node-wise transformations.
\newblock In {\em Proceedings of the IEEE/CVF Conference on Computer Vision and
  Pattern Recognition}, pages 7163--7172, 2020.

\bibitem{gidaris2020learning}
Spyros Gidaris, Andrei Bursuc, Nikos Komodakis, Patrick P{\'e}rez, and Matthieu
  Cord.
\newblock Learning representations by predicting bags of visual words.
\newblock In {\em Proceedings of the IEEE/CVF Conference on Computer Vision and
  Pattern Recognition}, pages 6928--6938, 2020.

\bibitem{gidaris2018unsupervised}
Spyros Gidaris, Praveer Singh, and Nikos Komodakis.
\newblock Unsupervised representation learning by predicting image rotations.
\newblock {\em arXiv preprint arXiv:1803.07728}, 2018.

\bibitem{goodfellow2014generative}
Ian Goodfellow, Jean Pouget-Abadie, Mehdi Mirza, Bing Xu, David Warde-Farley,
  Sherjil Ozair, Aaron Courville, and Yoshua Bengio.
\newblock Generative adversarial nets.
\newblock In {\em Advances in neural information processing systems}, pages
  2672--2680, 2014.

\bibitem{grill2020bootstrap}
Jean-Bastien Grill, Florian Strub, Florent Altch{\'e}, Corentin Tallec,
  Pierre~H Richemond, Elena Buchatskaya, Carl Doersch, Bernardo~Avila Pires,
  Zhaohan~Daniel Guo, Mohammad~Gheshlaghi Azar, et~al.
\newblock Bootstrap your own latent: A new approach to self-supervised
  learning.
\newblock {\em arXiv preprint arXiv:2006.07733}, 2020.

\bibitem{he2019momentum}
Kaiming He, Haoqi Fan, Yuxin Wu, Saining Xie, and Ross Girshick.
\newblock Momentum contrast for unsupervised visual representation learning.
\newblock {\em arXiv preprint arXiv:1911.05722}, 2019.

\bibitem{he2020momentum}
Kaiming He, Haoqi Fan, Yuxin Wu, Saining Xie, and Ross Girshick.
\newblock Momentum contrast for unsupervised visual representation learning.
\newblock In {\em Proceedings of the IEEE/CVF Conference on Computer Vision and
  Pattern Recognition}, pages 9729--9738, 2020.

\bibitem{henaff2019data}
Olivier~J H{\'e}naff, Aravind Srinivas, Jeffrey De~Fauw, Ali Razavi, Carl
  Doersch, SM Eslami, and Aaron van~den Oord.
\newblock Data-efficient image recognition with contrastive predictive coding.
\newblock {\em arXiv preprint arXiv:1905.09272}, 2019.

\bibitem{hjelm2018learning}
R~Devon Hjelm, Alex Fedorov, Samuel Lavoie-Marchildon, Karan Grewal, Phil
  Bachman, Adam Trischler, and Yoshua Bengio.
\newblock Learning deep representations by mutual information estimation and
  maximization.
\newblock {\em arXiv preprint arXiv:1808.06670}, 2018.

\bibitem{kalantidis2020hard}
Yannis Kalantidis, Mert~Bulent Sariyildiz, Noe Pion, Philippe Weinzaepfel, and
  Diane Larlus.
\newblock Hard negative mixing for contrastive learning.
\newblock {\em arXiv preprint arXiv:2010.01028}, 2020.

\bibitem{kim2018learning}
Dahun Kim, Donghyeon Cho, Donggeun Yoo, and In~So Kweon.
\newblock Learning image representations by completing damaged jigsaw puzzles.
\newblock In {\em 2018 IEEE Winter Conference on Applications of Computer
  Vision (WACV)}, pages 793--802. IEEE, 2018.

\bibitem{larsson2016learning}
Gustav Larsson, Michael Maire, and Gregory Shakhnarovich.
\newblock Learning representations for automatic colorization.
\newblock In {\em European conference on computer vision}, pages 577--593.
  Springer, 2016.

\bibitem{li2020prototypical}
Junnan Li, Pan Zhou, Caiming Xiong, Richard Socher, and Steven~CH Hoi.
\newblock Prototypical contrastive learning of unsupervised representations.
\newblock {\em arXiv preprint arXiv:2005.04966}, 2020.

\bibitem{lin2014microsoft}
Tsung-Yi Lin, Michael Maire, Serge Belongie, James Hays, Pietro Perona, Deva
  Ramanan, Piotr Doll{\'a}r, and C~Lawrence Zitnick.
\newblock Microsoft coco: Common objects in context.
\newblock In {\em European conference on computer vision}, pages 740--755.
  Springer, 2014.

\bibitem{loshchilov2016sgdr}
Ilya Loshchilov and Frank Hutter.
\newblock Sgdr: Stochastic gradient descent with warm restarts.
\newblock {\em arXiv preprint arXiv:1608.03983}, 2016.

\bibitem{misra2020self}
Ishan Misra and Laurens van~der Maaten.
\newblock Self-supervised learning of pretext-invariant representations.
\newblock In {\em Proceedings of the IEEE/CVF Conference on Computer Vision and
  Pattern Recognition}, pages 6707--6717, 2020.

\bibitem{oord2018representation}
Aaron van~den Oord, Yazhe Li, and Oriol Vinyals.
\newblock Representation learning with contrastive predictive coding.
\newblock {\em arXiv preprint arXiv:1807.03748}, 2018.

\bibitem{qi2016hierarchically}
Guo-Jun Qi.
\newblock Hierarchically gated deep networks for semantic segmentation.
\newblock In {\em Proceedings of the IEEE Conference on Computer Vision and
  Pattern Recognition}, pages 2267--2275, 2016.

\bibitem{qi2019learning}
Guo-Jun Qi.
\newblock Learning generalized transformation equivariant representations via
  autoencoding transformations.
\newblock {\em arXiv preprint arXiv:1906.08628}, 2019.

\bibitem{qi2020loss}
Guo-Jun Qi.
\newblock Loss-sensitive generative adversarial networks on lipschitz
  densities.
\newblock {\em International Journal of Computer Vision}, 128(5):1118--1140,
  2020.

\bibitem{qi2019small}
Guo-Jun Qi and Jiebo Luo.
\newblock Small data challenges in big data era: A survey of recent progress on
  unsupervised and semi-supervised methods.
\newblock {\em arXiv preprint arXiv:1903.11260}, 2019.

\bibitem{qi2019avt}
Guo-Jun Qi, Liheng Zhang, Chang~Wen Chen, and Qi Tian.
\newblock Avt: Unsupervised learning of transformation equivariant
  representations by autoencoding variational transformations.
\newblock In {\em Proceedings of the IEEE International Conference on Computer
  Vision}, pages 8130--8139, 2019.

\bibitem{qi2018global}
Guo-Jun Qi, Liheng Zhang, Hao Hu, Marzieh Edraki, Jingdong Wang, and Xian-Sheng
  Hua.
\newblock Global versus localized generative adversarial nets.
\newblock In {\em Proceedings of the IEEE conference on computer vision and
  pattern recognition}, pages 1517--1525, 2018.

\bibitem{tian2019contrastive}
Yonglong Tian, Dilip Krishnan, and Phillip Isola.
\newblock Contrastive multiview coding.
\newblock {\em arXiv preprint arXiv:1906.05849}, 2019.

\bibitem{wu2019detectron2}
Yuxin Wu, Alexander Kirillov, Francisco Massa, Wan-Yen Lo, and Ross Girshick.
\newblock Detectron2.
\newblock \url{https://github.com/facebookresearch/detectron2}, 2019.

\bibitem{wu2018unsupervised}
Zhirong Wu, Yuanjun Xiong, Stella~X Yu, and Dahua Lin.
\newblock Unsupervised feature learning via non-parametric instance
  discrimination.
\newblock In {\em Proceedings of the IEEE Conference on Computer Vision and
  Pattern Recognition}, pages 3733--3742, 2018.

\bibitem{zhang2019aet}
Liheng Zhang, Guo-Jun Qi, Liqiang Wang, and Jiebo Luo.
\newblock Aet vs. aed: Unsupervised representation learning by auto-encoding
  transformations rather than data.
\newblock {\em arXiv preprint arXiv:1901.04596}, 2019.

\bibitem{zhao2018adversarial}
Yiru Zhao, Zhongming Jin, Guo-jun Qi, Hongtao Lu, and Xian-sheng Hua.
\newblock An adversarial approach to hard triplet generation.
\newblock In {\em Proceedings of the European conference on computer vision
  (ECCV)}, pages 501--517, 2018.

\bibitem{zhuang2019local}
Chengxu Zhuang, Alex~Lin Zhai, and Daniel Yamins.
\newblock Local aggregation for unsupervised learning of visual embeddings.
\newblock In {\em Proceedings of the IEEE conference on Computer Vision}, 2019.

\end{thebibliography}
}

\clearpage
\appendix

\twocolumn
\twocolumn[
\section*{Appendix for ``AdCo: Adversarial Contrast for Efficient Learning of Unsupervised Representations from Self-Trained Negative Adversaries"}
]



In this appendix, we further analyze the impact of several factors on the performance of the learned representation.
We will demonstrate that
\begin{itemize}
\item Symmetrizing the contrastive loss has been proved effective in improving the performance of downstream tasks for existing models such as SimCLR \cite{chen2020simple} and BYOL \cite{grill2020bootstrap}. For a fair comparison, we also adopt it in AdCo, and show that its top-1 accuracy can be increased by $1.5\sim2.0\%$ that achieves the state-of-the-art result compared to its asymmetric counterpart.
\item AdCo is insensitive to the numbers of negative adversaries to pre-train the model. By reducing the number of negative samples from 65,536 to 8,192 (an eighth of the former size), AdCo has an unaffected top-1 accuracy. It shows that AdCo does not depend on a large amount of negative samples to pre-train the network.
\item We also attempt to answer an emerging question of if we still need the contrastive learning and the associated negative examples to pretrain a deep network while the BYOL \cite{grill2020bootstrap} does not rely on them any more. Specifically, the BYOL needs an extra MLP predictor with the same number of parameters as the negative examples that can be directly trained end-to-end in the AdCo. After comparing the computing costs among different methods, we show that the AdCo and its negative samples can be trained with almost 20\% less GPU time with a much smaller batch size than the BYOL and its MLP predictor.
\end{itemize}

\section{Symmetric Loss}\label{appendix:sloss}

Symmetrizing the loss in self-supervised learning has demonstrated improved performances for various models.  For example, in SimCLR,
a pair of positive samples augmented from the same image can exchange their roles in the contrastive loss  by viewing one as a query and the other as a key and vice versa.
Similarly, BYOL can also swap the inputs to its two branches, giving rise to a symmetric MSE term to minimize.  In both SimCLR and BYOL, the symmetric loss has successfully improved the performance of the learned representation in downstream tasks.

We show in Table~\ref{tab:sym_loss} that the performance of AdCo can also be improved by $1.5\sim2.0\%$ by symmetrizing its contrastive loss to update the representation network with the other factors (e.g., the number of negative samples and batch size) fixed. Thus, we will  apply the symmetric loss in the following study when comparing it with the other models.

For a fair comparison, all results in Table~\ref{tab:sym_loss} are obtained based on single-crop augmentations. We also tested the symmetric loss with multi-crop augmentations over 200 epochs of pre-training, but found that the symmetric loss only marginally improved the top-1 accuracy from $73.2\%$ to $73.6\%$. We hypothesize that  multi-crop augmentations have already leveraged multiple pairs of positive examples for the same image, and the benefit would vanish by symmetrizing the loss over those multi-crop augmentations.

\section{Numbers of Negative Adversaries}\label{appendix:nneg}
One of the most important factors that could impact the performance of the learned representation on downstream tasks is the number of negative samples. For the sake of a fair comparison with the other SOTA contrastive models particularly MoCo v2, we have fixed it to $65,536$ in experiments.  Here we will study if and how a smaller number of negative samples will impact the model performance.


The results in Table~\ref{tab:sym_loss} show that the top-1 accuracy of AdCo is almost unaffected when the number of negative samples is reduced from 65,536 to merely 8,192. This suggests that the AdCo is insensitive to the change in the size of negative samples, no matter whether the symmetric loss is applied. Indeed, the top-1 accuracy of AdCo only varies by $0.2\sim0.4$ as the negative samples decrease to an eighth of the original size.

We hypothesize that this is attributed to the efficient adversarial training that results in more informative negative samples to self-supervise the network in AdCo. In other words, even a smaller number of negative adversaries are sufficiently representative to cover the learned representation in the embedding space, and there is no need to bring in too many negative samples to learn a high-performing contrastive model.


\begin{table*}[!t]
\begin{center}
\caption{Top-1 accuracy under the linear evaluation on ImageNet with the ResNet-50 backbone. The table compares the methods over 200 epochs of pretraining with various batch sizes and numbers of negative samples. We also evaluate the impact of symmetric loss on these methods.
The results show that with a smaller batch size, AdCo achieves the same top-1 accuracy to BYOL, while the latter needs almost 20\% more GPU time to pretrain the model. AdCo also has a stable top-1 accuracy under different numbers of negative samples. All results are reported with a single-crop augmentation for a fair comparison.}
\label{tab:sym_loss}
\setlength{\tabcolsep}{3.5mm}{
\begin{tabular}{cccccc}
\toprule[1pt]
Method & Symmetric Loss & Batch Size & \#Neg. Samples & Top-1 Acc. & \begin{tabular}[c]{@{}l@{}}(GPU $\cdot$ Time)\\ /epoch\end{tabular}\\
\midrule[1pt]
SimCLR \cite{chen2020simple} & \checkmark & 8192 & -  & 67.0 & 1.92 \\
MoCo v2 \cite{chen2020improved}& & 256 & 65536  & 67.5 & 2.12 \\
BYOL \cite{grill2020bootstrap} & \checkmark & 4096 & -  & 70.6 & 4.10 \\
SimSiam \cite{chen2020exploring} & \checkmark & 256 & -  & 70.0 & -$^*$\\
\midrule
AdCo & & 256 & 65536 & 68.6 & 2.26 \\
AdCo &  & 256 & 16348  & 68.6 & 2.24 \\
AdCo & & 256 & 8192  & 68.4 & 2.24 \\
\midrule
AdCo & \checkmark & 256 & 65536 & 70.6 & 3.50 \\
AdCo & \checkmark & 256 & 16384 & 70.2 & 3.46 \\
AdCo & \checkmark & 256 & 8192 & 70.2 & 3.45 \\
\bottomrule[1pt]
\end{tabular}}
\end{center}
*Although no GPU time was reported on SimSiam, it should be on par with BYOL as its variant due to the similar network architecture and pre-training process.
\end{table*}

\section{Do We Still Need Negative Samples?}\label{appendix:nsamples}

MoCo v2 and BYOL are two state-of-the-art self-supervised models in literature. In Table~\ref{tab:sym_loss}, we compare the AdCo with them in terms of both accuracy and efficiency. The results show that both AdCo and BYOL outperform the compared models when the symmetric loss is applied.

Here, an insightful question may emerge -- while the BYOL removes the need of negative samples in self-training a deep network, do we still rely on the contrastive learning and the associated negative samples for network pretraining?

When we proceed to further compare BYOL and AdCo, we note that although BYOL does not explicitly use any negative samples, it depends on an extra prediction MLP to predict the embedding of an augmented view. There are empirical evidences \cite{chen2020exploring} showing that such a MLP predictor plays an indispensable role in obtaining competitive results in experiments.
Similarly, AdCo also has an extra network component by treating the trainable negative samples as an additional single neural layer attached onto the network to be pre-trained. In this sense, for a fair comparison, both BYOL and AdCo contain some extra model parameters. With the help of AdCo, it is now possible to directly train these parameters associated with negative samples end-to-end. While the number of parameters in the MLP predictor of BYOL is about 1M, the AdCo contains the same amount of trainable parameters associated with $8,192$ negative examples.

However, by comparison, BYOL needs almost 20\% more GPU time than AdCo to pre-train the network.  BYOL also relies on a larger batch size of $4,096$ than AdCo that has a much smaller batch size of 256 to achieve a competitive top-1 accuracy. This shows that AdCo can be trained in a more efficient fashion than BYOL.


\end{document}